\documentclass[conference]{IEEEtran}
\IEEEoverridecommandlockouts
\usepackage{cite}
\usepackage{amsmath,amssymb,amsfonts}
\usepackage{algorithmic}
\usepackage{graphicx}
\usepackage{textcomp}
\usepackage{xcolor}
\usepackage{newtxtext,newtxmath}

\usepackage{color}

\def\BibTeX{{\rm B\kern-.05em{\sc i\kern-.025em b}\kern-.08em
    T\kern-.1667em\lower.7ex\hbox{E}\kern-.125emX}}
\begin{document}

% \title{Conference Paper Title*\\
% \thanks{Identify applic_e funding agency here. If none, delete this.}
% }

\title{DreamScape: 3D Scene Creation via Gaussian Splatting joint Correlation Modeling}

% \author{
% 	Yueming Zhao*, Xuening Yuan* , Hongyu Yang$^{\dagger}$ ,Di Huang \\
% 	School of Computer Science and Engineering, \\ 
%     Beihang University, Beijing, China
% }
% \maketitle
% \renewcommand{\thefootnote}{}
% \footnotetext{* These authors contributed equally.}
% \footnotetext{$^{\dagger}$ Corresponding author: 3. Email: }

\author{
    Yueming Zhao$^{1}$*, Xuening Yuan$^{1}$*, Hongyu Yang$^{2,3}$$^{\dagger}$, Di Huang$^{1}$
\\
    $^{1}$School of Computer Science and Engineering, Beihang University, Beijing, China \\
    $^{2}$School of Artificial Intelligence, Beihang University, Beijing, China \\
    $^{3}$Shanghai Artificial Intelligence Laboratory, Shanghai, China\\
    \{zhaoyueming, snowing, hongyuyang, dhuang\}@buaa.edu.cn
}
\maketitle
\renewcommand{\thefootnote}{}
\footnotetext{* Both authors contributed equally to this rescarch.}
\footnotetext{$^{\dagger}$ Corresponding author: Hongyu Yang, hongyuyang@buaa.edu.cn}

\maketitle

\begin{abstract}
Recent advances in text-to-3D creation integrate the potent prior of Diffusion Models from text-to-image generation into 3D domain. Nevertheless, generating 3D scenes with multiple objects remains challenging. 
Therefore, we present DreamScape, a method for generating  3D scenes from text. Utilizing  Gaussian Splatting for 3D representation, DreamScape introduces 3D Gaussian Guide  that encodes semantic primitives, spatial transformations and relationships from text using LLMs, enabling local-to-global optimization.
Progressive scale control is tailored during local object generation, addressing training instability issue arising from simple blending in the global optimization stage. 
Collision relationships between objects are modeled at the global level to mitigate biases in LLMs priors, ensuring physical correctness. Additionally, to generate pervasive objects like rain and snow distributed extensively across the scene, we design specialized sparse initialization and densification strategy. Experiments demonstrate that DreamScape achieves state-of-the-art performance, enabling high-fidelity, controllable 3D scene generation.

\end{abstract}


\section{Introduction}
\label{sec:intro}
% Recent advancements in customized 3D content generation \cite{zero123,fantasia3d,sweetdreamer,syncdreamer} have achieved significant progress by leveraging the capabilities of large-scale pre-trained image generation models, such as Stable Diffusion \cite{stablediffusion}. These models have extended their remarkable generation ability to the 3D domain through the core concept of Score Distillation Sampling (SDS) \cite{dreamfusion}. By integrating priors such as mesh-based geometry constraints and point cloud diffusion, existing methods for 3D object generation \cite{shape,pointe} have demonstrated the capacity to synthesize corresponding 3D content solely from textual input, exhibiting commendable 3D coherence and high-fidelity details.
Recent progress in customized 3D content generation\cite{zero123} have achieved remarkable progress by leveraging the capabilities of large-scale pre-trained image generation models, such as Stable Diffusion\cite{stablediffusion}, extending their capabilities to the 3D domain via Score Distillation Sampling (SDS) \cite{dreamfusion}. Existing methods integrating priors like mesh geometry constraints and point cloud diffusion \cite{shape,pointe} can synthesize coherent and high-fidelity 3D objects solely  from textual inputs. 

\begin{figure}[ht]
\centering
\includegraphics[width=\linewidth]{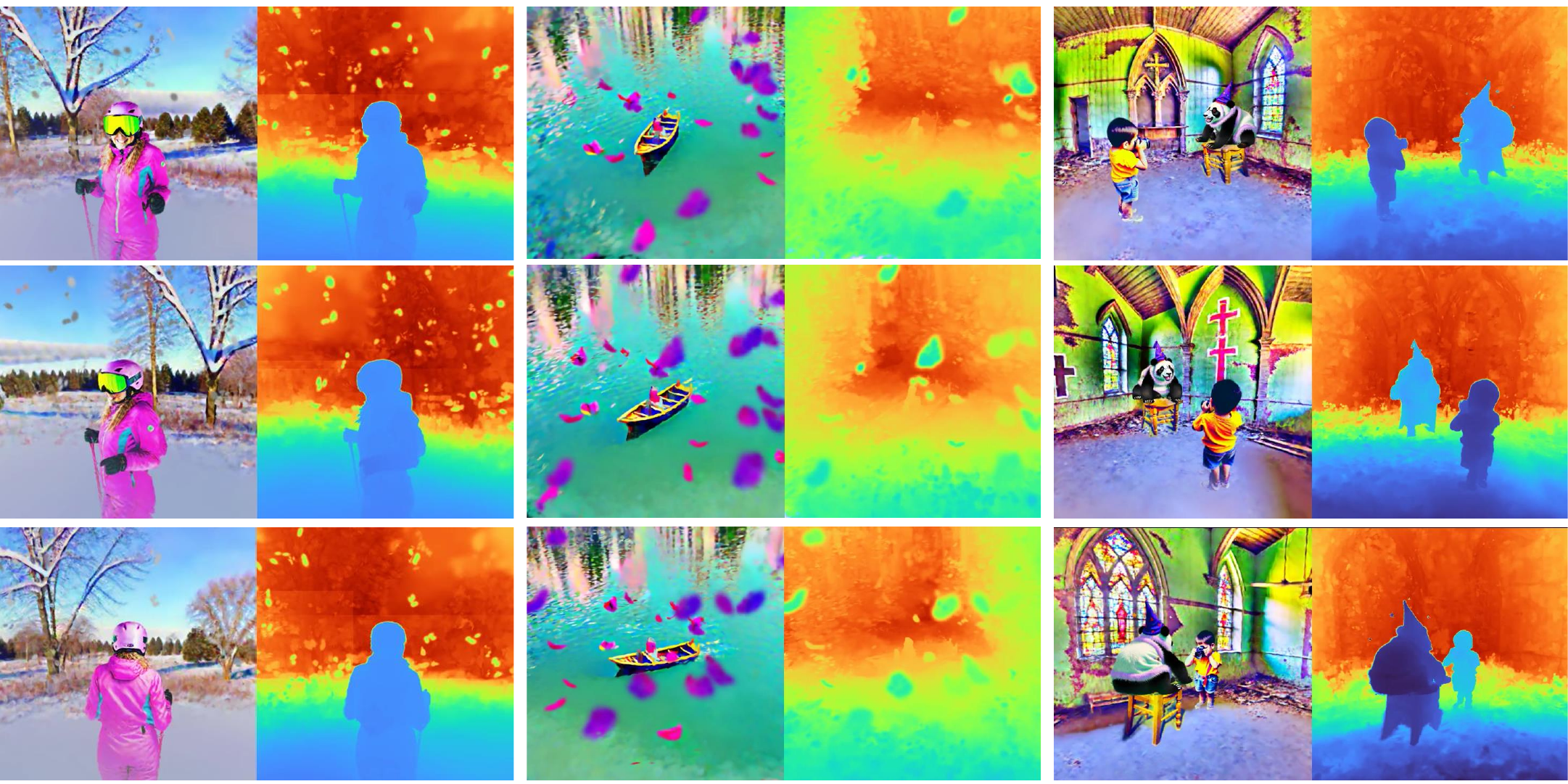} % Reduce the figure size so that it is slightly narrower than the column.
\caption{
DreamScape leverages Diffusion Models and LLMs to generate detailed, realistic, and multi-angle consistent scenes from text descriptions, demonstrating strong modeling across various scene types. This figure shows multi-view RGB images and depth maps generated by DreamScape.
%DreamScape combines the powerful prior in Diffusion Models and LLMs, enabling it to generate intricate details and realistic environments with multi-angle consistency based solely on textual descriptions, showcasing strong modeling capabilities across various scene types. This figure displays multi-view RGB images of scenes generated by the DreamScape alongside their corresponding depth maps. 
%\textbf{More cases and dynamic visualizations can be found in the supplementary materials.}
}
\label{fig2}
\end{figure}

However, the strategy of distilling 2D priors faces significant challenges when dealing with texts describing scenes with multiple objects. Current methods struggle with complex arrangements, leading to issues like textual guidance collapse, which fails to capture dense semantic concepts \cite{sjc,latentnerf,gsgen}, or poor generation quality, such as 3D inconsistencies and geometric distortions \cite{prolificdreamer}. Other approaches involve directly generating complex scene images using diffusion models, followed by in-painting and depth estimation techniques to elevate 2D contents into 3D representations \cite{text2immersion,luciddreamer}. However, these methods often lack genuine 3D information due to weak 3D lifting techniques. Consequently, the generated 3D results degrade significantly when the camera deviates from the training trajectory.

To address these challenges, several methods for text-to-scene generation \cite{sts,ctrlroom,graphdreamer,scenewiz3d} have been developed to explicitly model object arrangements in 3D space. These methods control the positions and transformations of objects through layout or positional proxies. However, they either require users to provide complex prompts \cite{sts,scenewiz3d}, which reduces the flexibility and efficiency of the generation process, or are limited by the drawbacks of their representations, such as NeRF \cite{nerf}, which lacks effective control mechanisms \cite{graphdreamer,sts} and high-frequency details. Recently, GALA3D \cite{gala3d} employs Gaussian Splatting \cite{3dgs} for efficient scene representation but fails to model background elements and object interactions.

% Recently, GALA3D \cite{gala3d} has shifted this paradigm to the 3D Gaussian \cite{3dgs} space, leveraging the efficient and strong representation ability of Gaussian Splatting. However, GALA3D does not address background modeling and the generation of pervasive objects, such as rain and snow, which are distributed extensively across the scene. These are critical characteristics that distinguish scene generation from single object generation and should be seriously considered.

In this paper, we introduce DreamScape, a novel approach for high-fidelity 3D scene generation from textual descriptions, leveraging Gaussian Splatting and Large Language Models (LLMs) to enhance fidelity and reduce discrepancies with textual descriptions.
Key to DreamScape is the use of a 3D Gaussian Guide , which serves as a comprehensive representation of the scene. This guide, derived from text prompts using LLMs, includes semantic primitives (objects), their spatial transformations, and scene correlations. 
It enables DreamScape to employ a local-global generation strategy, ensuring both instance-level realism and global consistency.

DreamScape employs a progressive scale guidance technique during local object generation. This technique considers the scale of each object in relation to the overall scene, allowing for more adaptive object generation. At the global level, DreamScape uses a collision loss between objects to prevent intersection and misalignment, addressing the potential spatial biases of $3{DG^2}$ provided by LLMs and ensuring physical correctness.
This dual-level optimization helps achieve instance-level realism and global consistency, enhancing interactions between objects such as water ripples, reflections, and lighting effects. 
To model pervasive objects like rain and snow, DreamScape introduces sparse initialization and it also incorporates densification and pruning strategies tailored to such objects, resulting in more realistic scenes.

Experimental results demonstrate its capability to faithfully generate 3D scenes from textual prompts while preserving semantic information. The approach achieves superior quality in 3D scene generation and supports various editing capabilities. 

The contributions of our paper are:
% \begin{itemize}
% \item We present DreamScape, a novel scene generation pipeline based on 3D Gaussian Splatting. The key component, $3{DG^2}$, effectively plans the entire scene, initializing scenes and facilitating subsequent local-global 3D Gaussian optimization process.
% \item A progressive scale constraint allows the model to adjust the scale proportions of objects while ensuring their appearance, thus avoiding distortion and stretching in the global optimization stage.
% \item DreamScape introduces the concept of pervasive objects, proposing sparse initialization and developing corresponding densification and pruning strategies for such objects.
% \end{itemize}
\begin{itemize} 
\item We introduce DreamScape, a 3D Gaussian Splatting-based pipeline, uses $3{DG^2}$ for scene planning, initialization, and facilitating local-global optimization. 
\item DreamScape proposes a local-global training strategy. In the local stage, object generation is guided by progressive scale constraints to preserve appearance and avoid distortion, while the global stage employs collision loss to maintain spatial alignment and physical correctness.
\item DreamScape introduces the concept of pervasive objects, proposing sparse initialization and developing corresponding densification and pruning strategies for such objects. 
\end{itemize}

\section{Related Work}
\subsection{3D Object Generation}

Existing methods for generating 3D objects from text are categorized into inference-based \cite{shape, pointe, mvdream, 3dgen, 3dgentriplane} and optimization-based approaches \cite{dreamfusion, magic3d, sjc, prolificdreamer, gaussiandreamer, luciddreamer, gsgen, dreamgaussian, latentnerf}. Inference-based methods, like Point-E \cite{pointe} and Shap-E \cite{shape}, generate 3D objects quickly using point cloud diffusion but require large, diverse datasets, potentially resulting in diminished geometric fidelity of the generated 3D objects. Optimization-based methods leverage pre-trained 2D diffusion models for 3D generation via Score Distillation Sampling (SDS), as seen in DreamFusion \cite{dreamfusion} and SJC \cite{sjc}, showcasing strong capabilities in 3D generation. Subsequent studies like Magic3D \cite{magic3d} refine 3D meshes through a coarse-to-fine approach, while Latent-NeRF \cite{latentnerf} integrates text and shape guidance using latent disentanglement.
The introduction of 3D Gaussian models \cite{3dgs} has further advanced the field. DreamGaussian \cite{dreamgaussian} and GSGEN \cite{gsgen} combine 3D Gaussians with geometry and texture optimization to improve 3D generation. However, these methods struggle with complex multi-object scenes. DreamScape addresses this limitation by modeling object-scene interactions for high-quality 3D scene generation.

\begin{figure*}[t]
  \centering
  \includegraphics[width=0.96\linewidth]{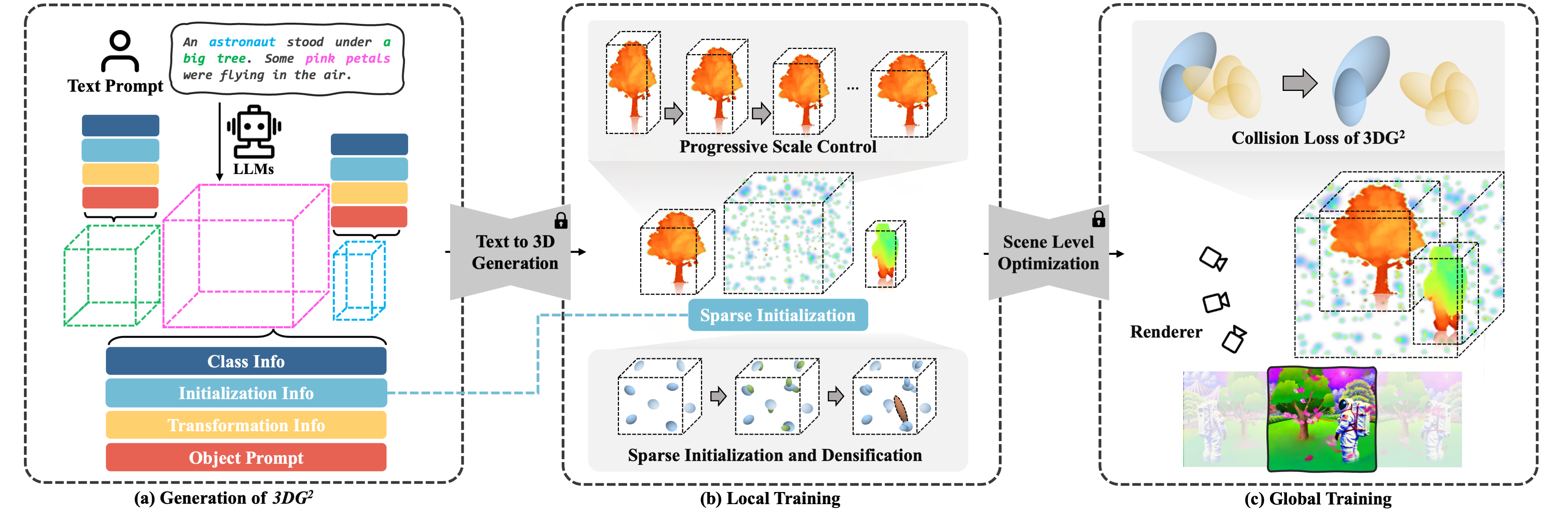} 
  % \caption{Overview of our method. Given a text prompt as input, DreamScape first generates $3{DG^2}$ corresponding to the text prompt using LLMs to help the model better understand the scene. DreamScape then undergoes local-global training with a frozen Diffusion Prior based on the $3{DG^2}$. During training, progressive scale control and synchronization optimization of $3{DG^2}$ are employed. Additionally, for pervasive objects, DreamScape utilizes special sparse initialization and densification strategies. The generated 3D content can be rendered from multiple views into coherent images.}
  
% \vspace{-3pt}
  % \caption{Overview of our method: (a) Given a text prompt, DreamScape utilizes $3{DG^2}$ generated by LLMs to interpret the scene, which then guides the local-global training process with a frozen diffusion prior. (b) The local step focuses on generating detailed individual objects, employing progressive scale control, sparse initialization and densification strategies. (c) In the global step, objects are aligned in a unified coordinate system via $3{DG^2}$ and refined based on viewing perspectives, ensuring detailed textures, consistent interactions, employing collision loss of $3{DG^2}$.}

  \caption{Overview of our method: (a) Given a text prompt, DreamScape uses $3{DG^2}$ generated by LLMs to interpret the scene, guiding local-global training with a frozen diffusion prior. (b) The local step generates detailed objects using progressive scale control, and sparse initialization and densification for pervasive objects. (c) In the global step, objects are aligned in a unified coordinate system via $3{DG^2}$, refined based on perspectives, ensuring detailed textures and consistent interactions with collision loss of $3{DG^2}$.}

  \label{fi:pipeline}
\end{figure*}
\subsection{3D Scene Generation}

% Current 3D scene generation approaches encounter notable constraints in producing high-quality and controllable 3D scenes. Text2-Room \cite{text2room}  generates textured 3D meshes depicting room-scale scenes from textual prompts. Text2NeRF \cite{text2nerf} combines diffusion models and NeRF representations, enabling zero-shot generation of diverse indoor and outdoor scenes. Despite their proficiency in generating the geometry of entire rooms, these scene generation method based on image inpainting \cite{text2room,text2nerf, text2immersion, ctrlroom} exhibit deficiencies in individual object modeling and exhibit limited 3D consistency.

Current 3D scene generation methods struggle with producing high-quality, controllable scenes. Text2Room \cite{text2room} generates textured room-scale meshes from text, while Text2NeRF \cite{text2nerf} combines diffusion models with NeRF to create diverse indoor and outdoor scenes. However, these methods, often based on image inpainting \cite{text2room, text2nerf, text2immersion, ctrlroom}, lack 3D consistency and struggle with individual object modeling.
Object-centric approaches \cite{componerf, cg3d, gala3d, comp3d, instructscene, showroom3d} generate complex scenes by combining objects but lack global constraints, hindering object interactions. Set-the-Scene\cite{sts} uses a proxy-based training framework to create harmonious scenes, but requires complex constraints and lacks flexible editing. GALA3D \cite{gala3d} introduces layout guidance generated by LLMs but lacks modeling background and object interactions. LucidDreamer \cite{luciddreamer} generates 3D-consistent scenes from point clouds albeit with limited perspectives. Moreover, most methods fail to model pervasive objects like snowflakes, limiting scene realism. 
In contrast, DreamScape excels in generating interactive 3D scenes and introduces \textit{pervasive object} modeling, enhancing its capability to handle diverse scenarios.

\section{Method}

Figure \ref{fi:pipeline} illustrates DreamScape. Beginning with a textual input, DreamScape utilizes LLMs to parse the scene and generate the initial 3D Gaussian Guide  of the target scene (Section \ref{ss:MOR}). This guide comprises semantic primitives, spatial transformations and scene correlations, providing a foundational representation of the scene. DreamScape then initializes Gaussians for each object and employs a local-global training strategy to refine the 3D representations. During local optimization, a progressive scale control ensures alignment, while global optimization of the entire $3{DG^2}$ is performed to achieve overall scene consistency. To enhance realism, DreamScape introduces a collision training loss (Section \ref{ss:sync}). Additionally, for pervasive objects like rain and snow, DreamScape employs a sparse initialization method, along with densification and pruning operations, to effectively model such objects (Section \ref{ss:spa}).

\subsection{3D Gaussian Guide}
\label{ss:MOR}

% Due to the ambiguous nature of textual prompts, current methods for 3D scene generation often struggle to balance convenience and controllability. They typically rely on intricate shape control \cite{sts, comp3d} or face challenges in generating controllable scenes \cite{prolificdreamer, luciddreamer} that accurately match the given descriptions. Recently, methods like GALA3D \cite{gala3d}, GraphDreamer \cite{graphdreamer}, and SceneWiz3D \cite{scenewiz3d} have leveraged LLMs to provide prior information about object positions in scenes, yielding promising results. Similarly, we introduce LLMs to offer additional priors for scenes, enabling us to acquire more information than relying solely on the diffusion model without increasing user input.

Due to the ambiguity of textual prompts, current 3D scene generation methods face challenges in balancing convenience and controllability, often requiring complex shape controls \cite{sts, comp3d} or struggling to generate scenes matching descriptions \cite{prolificdreamer, luciddreamer}. Recent methods like GALA3D \cite{gala3d}, GraphDreamer \cite{graphdreamer}, and SceneWiz3D \cite{scenewiz3d} use LLMs to provide object position priors, achieving notable improvements. Similarly, we leverage LLMs to supply additional priors, enhancing scene generation without increasing user effort.

To ensure usability and controllability, $3{DG^2}$ serves as a scene framework, translating textual prompts into a guide for 3D Gaussian scene generation. Leveraging LLMs, it encodes object distributions and spatial transformations for initialization and optimization. Specifically, $3{DG^2}$ consists of the following parameters:

\begin{small}
\begin{equation}
    3{DG^2} = \{(cls_i, init_i, trans_i, prompt_i), i\in [1, \dots, N]\},
\end{equation}
\end{small}

\noindent where $cls_i$ denote the category of the $i$-th object, $init_i$ represent its initialization details (e.g., method, number, and color of points), $trans_i = (\mathbf{xyz}, \mathbf{whl}, \mathbf{quad})$ includes the object's position $\mathbf{xyz} \in \mathbb{R}^3$, scale $\mathbf{whl} \in \mathbb{R}^3$, and rotation $\mathbf{quad} \in \mathbb{R}^4$. $prompt_i$ is the textual description. These parameters, generated by LLMs, are refined during optimization.
% where $cls_i$ indicates the category information of the $i$-th object, $init_i$ is the initialization information of this object, including the initialization method (by either Point-E or sparse initialization), number and color of the initialized points, \textit{etc.}. $trans_i$ is a tuple of the form $(\mathbf{xyz},\mathbf{whl}, \mathbf{quad})$, where $\mathbf{xyz} \in \mathbb{R}^{3}$ indicates the position of the object center in the scene coordinate system, $\mathbf{whl} \in \mathbb{R}^{3}$ represents the scale of an object, and $\mathbf{quad} \in \mathbb{R}^{4}$ is a quadruple representing the rotation of an object; $prompt_i$ is the detailed textual description. All of these parameters can be directly generated by LLMs, and the parameters $trans_i$ will be further refined during model optimization.

We uses 3D Gaussians for scene representation, formulated as:

\vspace{-0.2cm}
\begin{small}
\begin{equation}
 O_i = ( \mathbf{p}, \mathbf{s}, \mathbf{q}, \mathbf{c}, \alpha),
\end{equation}
\begin{equation}
 S = \{ 3{DG^2}, O_i, i \in [1, \dots, N ] \},
\end{equation}
\end{small}

\noindent where $\alpha$ is the opacity, $\mathbf{p}, \mathbf{s},\mathbf{c} \in \mathbb{R}^{N\times 3}$ and $\mathbf{q} \in \mathbb{R}^{N\times 4}$ denote the vectors of center position, scale matrix, color and rotation quadruple, as we convert the covariance of the 3D Gaussians into scale matrix and rotation quadruple for easier optimization. A set of Gaussians forms a 3D Gaussian object, and a collection of Gaussian objects along with the $3{DG^2}$ of the scene constitute a 3D Gaussian scene.

% In the local step, the center of an object is located at the center of the rendered area; in the global step, the coordinates need to be converted according to the location arrangements in the $3{DG^2}$. During the process of transforming the object from its local coordinate system to the scene coordinate system, the following formula can be used to obtain its new position, rotation, and scale information within the scene: 

In the local step, an object's center is at the rendered area's center. In the global step, coordinates are adjusted based on $3{DG^2}$. The following formula is used to transform the object from its local to scene coordinate system, yielding its new position, rotation, and scale:

\begin{small}
\begin{equation}
    \mathbf{p}'=\mathbf{p}*T_{\phi}\left[E_{\phi}(\mathbf{quad_i}) \hat{\times} \mathbf{q} \right]+E_{\phi}(\mathbf{xyz_i}), \label{eq:xyz_trans}
\end{equation}
\begin{equation}
    \mathbf{s}'=\mathbf{s}\cdot E_{\phi}(\mathbf{whl_i}), \label{eq:svec_trans}
\end{equation}
\begin{equation}
    \mathbf{q}'=E_{\phi}(\mathbf{quad_i}) \hat{\times} \mathbf{q},
    \label{eq:qvec_trans}
\end{equation}
\end{small}

\noindent where $T_{\phi}$ denotes the transformation from quadruple to rotation matrix, $\hat{\times}$ denotes the non-commutative quaternion multiplication, $E_{\phi}$ is the function that extends a vector from the first dimension to a certain length.
Similarly, objects can be restored to their original views through the inverse of these transformations. The alpha and color properties of Gaussian points do not require transformation.

\subsection{Scene Optimization} \label{ss:sync}

% \textbf{Local-global training strategy.} Due to the dense semantic concepts and complex colors and geometries, directly distilling the diffusion prior for the entire scene is impractical. Therefore, we adopt a dual-level training strategy for improved results. At the local level, we focus on generating individual objects to enhance details for high fidelity. Then, we collaboratively optimize the entire scene through global steps, to enhance global consistency and capture interactions among objects, rendering effects such as water ripples, reflections, and coordinated lighting. The position conversion of Gaussians between local and global steps can be referred to in formulas \ref{eq:xyz_trans}, \ref{eq:svec_trans}, and \ref{eq:qvec_trans}.

\textbf{Local-global training strategy.} Due to the complexity of semantic concepts, colors, and geometries, directly distilling the diffusion prior for the entire scene is impractical. Therefore, we use a dual-level training strategy: local steps focus on generating detailed individual objects, while global steps optimize the entire scene for consistency and object interactions, such as water ripples, reflections, and lighting effects. Position conversion of Gaussians between local and global steps is described in formulas \ref{eq:xyz_trans}, \ref{eq:svec_trans}, and \ref{eq:qvec_trans}.

Inspired by recent studies, DreamScape adopts SDS loss for optimizing 3D content from 2D diffusion priors:

\begin{small}
\begin{equation}
    \mathcal{L_{\text{SDS}}} = 
    \mathbb{E}_{\epsilon, t}\left[\omega(t)(\epsilon_\phi(x_t; y, t) - \epsilon_t)\frac{\partial x}{\partial \theta}\right],
\end{equation}
\end{small}

\noindent where $\epsilon_t$ is Gaussian noise at timestep $t$, and $\epsilon_\phi(x_t; y, t)$ represents the noise predicted by the pre-trained diffusion model, with $x_t$ and $y$ denoting the noisy image and textual prompt embedding, respectively. 
% The rendering of $x$ follows:
% \begin{equation}
%     x(p_x,p_y)=\sum_{i \in \mathcal{N}}{c_i\alpha_i}\prod_{j = 1}^{i-1}(1-\alpha_j). \label{eq:render}
% \end{equation}

In the local steps, DreamScape optimizes each object for a 360-degree panoramic view, ensuring 3D consistency. In the global step, objects are transformed into a unified coordinate system via $3{DG^2}$ and refined based on viewing perspectives, achieving detailed textures and consistent interactions among objects.

% \textbf{Progressive scale control.} In order to align objects with $3{DG^2}$, the model stretches objects along all the dimensions for further blending. However, if stretching occurs too early, the object may lose its initial geometric shape, which is detrimental to maintaining the 3D consistency of the object. Conversely, if stretching occurs after the object generation is completed, it may result in distorted textures and geometry, leading to a significant decrease in generation quality. Therefore, we propose progressive scale control, gradually increasing the influence of scale conditions on the appearance of objects during the object generation process, formulated as:
\textbf{Progressive scale control.} In order to align objects with $3{DG^2}$, the model stretches them across all dimensions for blending. However, premature stretching may distort the object's initial geometric shape, harming 3D consistency, while stretching after generation completed may distort textures and geometry, reducing generation quality. To address this, we propose progressive scale control, gradually increasing the scale influence during object generation, formulated as:

% \begin{equation}
%     \beta = \mathbf{whl} \cdot \left[max(\textbf{xyz})-min(\textbf{xyz})\right]^{-1},
% \end{equation}

\begin{small}
\begin{equation}
\begin{bmatrix}
\hat{\textbf{p}} \
\hat{\textbf{s}}
\end{bmatrix}
=
\begin{bmatrix}
\textbf{p} \
\textbf{s}
\end{bmatrix}
\cdot E_\phi(\textbf{I} + \beta \cdot min(\text{ReLU}(\frac{k-w}{\gamma}), 1)),
\end{equation}
\end{small}

\noindent where $\beta= \mathbf{whl} \cdot \mathbf{s}^{-1} - 1$ is the scale factor, 
$k$ denote the number of training steps completed for each object, $w$ represents the number of warm-up steps for the object, 
$\gamma$ denotes the saturation step for scale control.
With progressive scale control, objects can gradually converge to the scale provided by $3{DG^2}$ while maintaining good geometric shapes and texture features.

% \textbf{Synchronized optimization of $3{DG^2}$.} Despite the remarkable understanding capability of current LLMs, there is still a possibility of providing incorrect priors. LLMs may yield conflicting object positions, leading to the phenomenon of object intersection. Therefore, DreamScape sets the $3{DG^2}$ as optimizable parameters. During global training, specific object position and scale information will be optimized for $3{DG^2}$. The corresponding loss function is defined as follows:

\textbf{Synchronized optimization of $3{DG^2}$.} Despite the strong understanding capabilities of LLMs, they may provide incorrect priors, causing conflicting object positions. To address this, DreamScape treats $3{DG^2}$ as optimizable parameters, refining object position and scale during global training. The corresponding loss function is defined as:

\begin{small}
\begin{equation}
    \mathcal{L_{\text{cross}}} = \mathcal{C}_\phi(p_i,p_j,\theta), \quad i,j \in [1,\dots,N].
\end{equation}
\end{small}
% We define a simple function $\mathcal{C}_\phi$ as a representation of collisions between objects. This function queries the sum of the distance between points that are closer to each other among two objects and filters based on a threshold value $\theta$. DreamScape efficiently implements this functionality using KD-trees \cite{kdtree1, kdtree2}, avoiding complex computational processes when querying collision situations. Under the constraints of this function, the initialization of collision positions in $3{DG^2}$ will be optimized, thereby avoiding instances of objects crossing each other. In particular, due to the particularity of pervasive objects, we do not calculate collision loss for such objects.

\noindent where $\mathcal{C}_\phi$ is a collision function that calculates the sum of distances between points from two objects that are close, filtering based on a threshold $\theta$. DreamScape efficiently implements this using KD-trees \cite{kdtree1, kdtree2}, reducing computational complexity. This function helps optimize collision positions in $3{DG^2}$ to prevent object intersections, with the exception of pervasive objects, for which collision loss is not computed.

The overall training loss of our method is:

\vspace{-0.2cm}
\begin{small}
\begin{equation}
    \mathcal{L}=\lambda_{1}\sum_{i=1}^{N}\mathcal{L}_{SDS\_local_i} + \lambda_{2}\mathcal{L}_{cross} + \lambda_{3}\mathcal{L}_{SDS\_global},
\end{equation}
\end{small}

\noindent where $\mathcal{L}_{SDS\_local}$, $\mathcal{L}_{SDS\_global}$ are the losses of score distillation in the local and global steps, respectively, and $\mathcal{L}_{cross}$ is the collision loss of $3{DG^2}$.

\subsection{Sparse Initialization and Densification} \label{ss:spa}

% Due to the characteristics of 3D Gaussians in representing objects, existing 3D content generation methods \cite{gsgen, dreamgaussian, gaussiandreamer} tend to produce dense, surface-floating Gaussians to achieve optimal detail representation. This strategy is not favorable for objects composed of numerous sparse small elements, which would quickly cause sparse objects to stick together. However, these pervasive objects are important for scene composition in some special conditions, including generating snow scenes, floating small petals, and so on. Using multiple objects to represent a pervasive object is undoubtedly resource-wasting. Therefore, DreamScape introduces the concept of \textit{pervasive object} to represent objects composed of numerous sparse small elements. For pervasive objects, we proposes sparse initialization and sparse densification strategies to optimize performance. 
Existing 3D generation methods \cite{gsgen, dreamgaussian, gaussiandreamer} use dense, surface-floating Gaussians for optimal detail, which is not ideal for sparse objects, as they may quickly cluster together. However, pervasive objects, such as snow or floating petals, require special handling. Instead of using multiple objects, DreamScape introduces \textit{pervasive objects} to represent such elements, applying sparse initialization and densification strategies to optimize performance.

\textbf{Sparse initialization.} 
% DreamScape randomly samples a small number of points within the bounding box of pervasive objects for initialization, similar to a \textit{condensation nucleus}, around which the Gaussians of pervasive objects are subsequently densified. Since Gaussian points will move during subsequent optimization processes, a viable setup is to use uniform sampling, formulated as:
% \begin{equation}
%     \mathbf{x} \sim \text{Uniform}(a, b)^3,
%     \label{eq: unifo}
% \end{equation}
% where $\mathbf{x} \in \mathbb{R}^3$ represents the position of a Gaussian point. with each element independently sampled from a uniform distribution over the interval $\left[a,b\right]$. 
DreamScape initializes pervasive objects by randomly sampling a small number of points within their bounding box, serving as “condensation nuclei" for subsequent Gaussian densification. To ensure flexibility during optimization, the initialization uses a uniform distribution to determine the positions of these points, allowing for even dispersion across the bounding box.

\begin{figure*}[htb]
  \centering
  \includegraphics[width=0.90\linewidth]{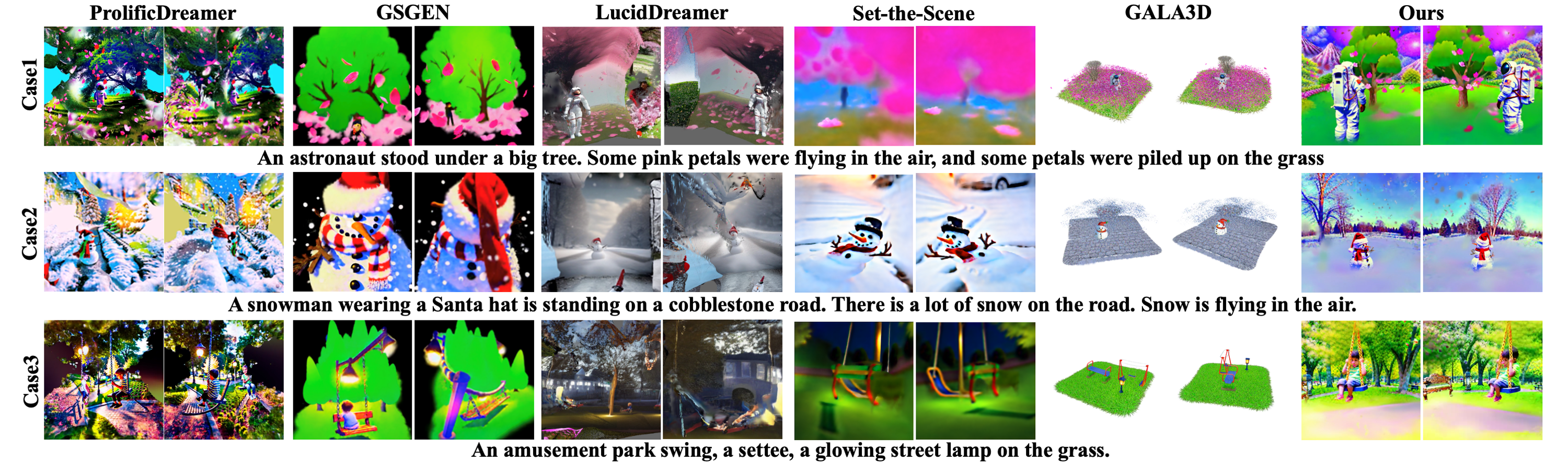}
  % \caption{Qualitative comparisons of typical text-to-3D generation methods (zoom in for a better view).}
  \caption{Qualitative comparisons with state-of-the-art text-to-3D generation methods.}
  \label{fi:comparisons} 
\end{figure*}

\textbf{Sparse densification and pruning strategy.} For conventional objects, densification prioritizes filling holes with high-frequency Gaussians, while pruning removes isolated ones. This approach, however, is unsuitable for pervasive objects. DreamScape adjusts the strategy by reducing densification frequency and pruning Gaussians with large scales to avoid clustering. With sparse initialization, this method effectively prevents oversized Gaussian clusters in pervasive objects and optimizes the appearance of small elements at the scene scale. We set the adjusted frequency for densification $\nu ' = \tau \cdot \nu $, where \( \nu\) is the original densification frequency, and \( \tau \) is the adjustment factor. For the pruning strategy, a threshold \( \rho_{\theta} \) is set. Gaussians greater than \(\rho_{\theta} \) will be preferentially removed during the pruning process.
%For conventional objects, the densification process typically prioritizes areas with holes for high-frequency densification, while adopting a pruning strategy for isolated Gaussians. However, this strategy is unsuitable for pervasive objects. DreamScape modifies the strategy for pervasive objects. Specifically, the frequency of densification are appropriately reduced, and the pruning strategy considers pruning Gaussians with large scales to prevent object clustering. Under the premise of sparse initialization, this strategy simply and effectively prevents the generation of overly large Gaussian clusters within pervasive objects and allows for appearance optimization of small objects at the scene scale.

% Sparse densification can be formulated as:
% \begin{equation}
%     \nu ' = \tau \cdot \nu ,
% \end{equation}
% where \( \nu\) is the original densification frequency, \( \nu ' \) indicates the adjusted frequency, and \( \tau \) is the adjustment factor. For the pruning strategy, a threshold \( \rho_{\theta} \) is set. Gaussians greater than \(\rho_{\theta} \) will be preferentially removed during the pruning process.

\section{Experiments}

We employ GSGEN \cite{gsgen} as the 3D content generation baseline, which exhibits decent performance on the generation of 3D Gaussian objects. Implementation details can be found in supplementary materials.

\subsection{Quantitative Comparison}

% In order to evaluate the capabilities of our model, we have conducted a comprehensive comparison with existing state-of-the-art generative models, including single-object generation methods \cite{sjc,latentnerf,gsgen,prolificdreamer}, 2D to 3D lifting method \cite{luciddreamer} and scene generation method \cite{sts}. Specifically, Set-the-Scene requires precise object shape proxies to control geometries and positions of the objects, so we directly adopt the simplified positional proxies provided by LLMs for scene representation. 
% CLIP similarity is used to measure semantic accuracy between the generated results of the models and the original text prompts. We evaluate on 3 cases and captured 10 views in the rendered results of 3D content as image outputs to measure the CLIP similarity with the input texts. The comparative results are presented in Table \ref{tab:clipsim}.

We evaluate our model against state-of-the-art methods,  including single-object generation methods, \textit{i.e. } GSGEN \cite{gsgen} and ProlificDreamer \cite{prolificdreamer}, 2D to 3D lifting method, \textit{i.e. } LucidDreamer \cite{luciddreamer}, and scene generation methods, namely Set-the-Scene\cite{sts} and GALA3D \cite{gala3d}. For Set-the-Scene and GALA3D, simplified positional proxies from LLMs are used. CLIP similarity is used to measure semantic accuracy between generated results and text prompts. We evaluate 3 cases, capturing 10 views from the rendered 3D content to measure CLIP similarity, with results presented in Table \ref{tab:clipsim}.

\begin{table}[t]
\caption{CLIP similarity comparison with existing state-of-the-art text-to-3D methods.}
\begin{center}
\begin{tabular}{|c|c|c|c|c|}
\hline
\textbf{Method} & \textbf{Case1} & \textbf{Case2} & \textbf{Case3} & \textbf{Average} \\
\hline
% SJC & 0.27 & 0.19 & 0.28 & 0.26 \\
% \hline
% LatentNeRF & 0.30 & 0.22 & 0.31 & 0.28 \\
% \hline
ProlificDreamer & 0.32 & 0.25 & 0.31 & 0.29 \\
\hline
GSGEN & 0.29 & \textbf{0.29} & 0.30 & 0.29 \\
\hline
LucidDreamer & \textbf{0.35} & 0.28 & 0.32 & 0.31 \\
\hline
Set-the-Scene & 0.27 & 0.27 & 0.33 & 0.29 \\
\hline
GALA3D & 0.32 & 0.28 & 0.33 & 0.31 \\
\hline
Ours & 0.34 & \textbf{0.29} & \textbf{0.34} & \textbf{0.32} \\
\hline
\end{tabular}
\label{tab:clipsim}
\end{center}
\end{table}

% \textbf{Comparison with object generation methods.} Methods targeting single-object generation include SJC \cite{sjc}, LatentNeRF \cite{latentnerf}, ProlificDreamer \cite{prolificdreamer}, GSGEN \cite{gsgen}. These approaches often suffer from semantic loss of certain aspects of the text prompts during the scene generation due to their modeling paradigm, particularly when dealing with more complex scenes.
% In contrast, DreamScape employs a process of decomposing the objects within the text prompt, with each object individually modeled, substantially enhancing the fidelity to the text prompts, as evidenced by a significant improvement in average CLIP similarity.
\textbf{Comparison with object generation methods.} Single-object generation methods like ProlificDreamer \cite{prolificdreamer} and GSGEN \cite{gsgen} often experience semantic loss when generating complex scenes. In contrast, DreamScape decomposes objects in the text prompt, modeling each object individually, which significantly improves fidelity to the prompts, as shown by a notable increase in average CLIP similarity.

% \textbf{Comparison with scene generation methods.} Methods targeting scene generation mainly include Set-the-Scene \cite{sts} and LucidDreamer \cite{luciddreamer}.  
% Set-the-Scene is effective in modeling objects within scenes due to its precise positional control, but its generative performance is hindered by the lack of detailed shape proxies. LucidDreamer, which uses images as prompts, achieves relatively high average CLIP similarity. However, it is constrained to generating satisfactory results only from a limited range of viewpoints close to the provided image guidance.
% In contrast to these methods, DreamScape requires only a textual prompt as input to yield consistent results across various views, without being limited by perspective. This approach faithfully reflects the descriptions provided in the text, demonstrating its robustness in 3D scene generation.
% \textbf{Comparison with scene generation methods.} Scene generation methods like Set-the-Scene \cite{sts}, LucidDreamer \cite{luciddreamer} and GALA3D \cite{gala3d} face limitations. Set-the-Scene excels at modeling objects but struggles with detailed shape proxies, while LucidDreamer, using images as prompts, performs well from limited viewpoints. GALA3D is proficient in object combination but is limited in  background modeling and object interactions, resulting in less cohesive scene composition. In contrast, DreamScape only requires textual input, generating consistent results across various views, without perspective limitations, and faithfully reflects the text descriptions.
\textbf{Comparison with scene generation methods.}  Set-the-Scene\cite{sts} offers precise positional control but lacks detailed shape proxies, limiting generative performance. LucidDreamer \cite{luciddreamer} achieves high CLIP similarity using image prompts but is constrained to specific viewpoints. GALA3D\cite{gala3d} combines objects effectively but struggles with background modeling and object interactions, leading to less cohesive scenes. In contrast, DreamScape can producing consistent, perspective-free results that accurately reflect descriptions.

\subsection{Qualitative Comparison}

% We selected 3 representative scene generation tasks aligning with common practices\cite{luciddreamer,gala3d}, ensuring that the experimental settings remained consistent with the official documentation of the contrasted methods. 
% As previously mentioned, for Set-the-Scene \cite{sts} and LucidDreamer \cite{luciddreamer}, we also implemented reasonable additional generation conditions to ensure maximum fairness in the experiments. 
We display a qualitative comparison with the state-of-the-art generative methods to further demonstrate the capabilities of our model.Figure \ref{fi:comparisons} illustrates the comparative results of our model against existing generative methods. The text prompts used for generation are displayed below each row of images for clarity. Additional results can be found in the supplementary materials.
%more results can be seen in supplementary materials.

% \begin{table}[t]
% \centering

% \begin{tabular}{cccccc}
% \toprule
% Method & Case1 & Case2 & Case3 & Ave. \\
% \midrule
% SJC& 0.27 & 0.19 & 0.28 & 0.26 \\
% LatentNeRF & 0.30 & 0.22 & 0.34 & 0.29 \\
% LucidDreamer& \textbf{0.35} & 0.28 & 0.31 & 0.31 \\
% ProlificDreamer & 0.32 & 0.25 & 0.31 & 0.29\\
% GSGEN& 0.29 & 0.29 & 0.30 &0.29\\
% Set-the-Scene & 0.27 & 0.27 & 0.33 & 0.29 \\
% Ours & 0.34 & \textbf{0.29} & \textbf{0.31} & \textbf{0.32}\\

% \bottomrule
% \end{tabular}
% \caption{CLIP similarity comparison with existing state-of-the-art text-to-3D methods.}
% \label{tab:clipsim}
% \end{table}

\textbf{Comparison with object generation methods.} ProlificDreamer \cite{prolificdreamer} excels in scene context adaptability but struggles with multi-object modeling, causing inconsistencies and incomplete renderings from some angles. GSGEN \cite{gsgen} generates “multi-faces" scenes, offering multiple viewpoints of a single object’s outer surface. While effective for individual objects, these methods fall short in scene generation and do not match the performance of our approach.
% When compared to methods focused on single-object generation, the limitations of these approaches in scene generation become apparent, as illustrated in the figure.
% ProlificDreamer \cite{prolificdreamer}, while showing good adaptability to scene contexts, can only partially model scenes. It struggles with comprehensive multi-object modeling, leading to inconsistencies across different viewpoints and resulting in broken or incomplete renderings from certain angles. Similarly, GSGEN \cite{gsgen} often generates ``multi-faces" scenes, creating the appearance of a scene from various angles but only on the outer surface of a single object.
% Although these methods are effective in generating individual objects, they fall short in scene generation and do not match the performance of our approach.

\textbf{Comparison with scene generation methods.} Scene generation methods generally produce plausible spatial relationships. However, Set-the-Scene \cite{sts} produces blurry outputs despite extensive training and object proxies from LLMs. LucidDreamer \cite{luciddreamer} performs well but is limited to narrow angles, with unpredictable results for wider views. GALA3D \cite{gala3d} ensures high scene consistency but underutilizes LLMs, leading to rigid scenes and difficulties modeling backgrounds, interactions, or pervasive effects like rain and snow. DreamScape, in contrast, generates stable, consistent results from any perspective with a single text prompt.

\begin{table}[htbp]
\caption{User study results evaluating the quality and consistency of ablation studies}
\begin{center}
\begin{tabular}{|c|c|c|}
\hline
\centering \textbf{Method} & \textbf{Quality} & \textbf{Consistency}\\
\hline
\centering w/o Local-global training & 3.316 & 3.467  \\
\hline
%with Synchronized optimization of 3DG^2 & 3.716 & 3.652 & 3.731   \\
\centering w/o Synchronized optimization & 2.804 & 3.546  \\
\hline
%with Progressive scale control & 3.641 & 3.587 & 3.701   \\
\centering w/o Progressive scale control & 2.817 & 3.536  \\
\hline
%with Sparse initialization and densification & 3.714 & 3.785 & 3.821   \\
\centering w/o Sparse initialization and densification & 1.513 & 2.619  \\
\hline
\centering Full Model & \textbf{3.676} & \textbf{3.651} \\
\hline
\end{tabular}
\label{tab:userstudy}
\end{center}
\end{table}

\subsection{Ablation Studies}

% In order to validate the effectiveness of the proposed modules, we conducted ablation experiments on the local-global training, progressive scale control, synchronized optimization and sparse initialization and densification methods. We visually compared the results of the ablation experiments, as shown in Figure \ref{fi:ablation}. 

% It can be observed that without local-global training, the model lacks modeling of interactions between objects, whereas with local-global training, reflection of the boat will be on the lake. Objects may collide with each other without optimizing $3{DG^2}$, which is well solved by synchronized optimization of $3{DG^2}$. Objects may undergo stretching in the global step, resulting in abnormal deformation. And with the addition of progressive scale control, objects can gradually deform to the required scale for the scene. By introducing sparse initialization and densification, pervasive objects can be accurately modeled. The experiments demonstrate that our proposed optimization modules are highly effective in enabling the model to generate higher-quality results.

\begin{figure}[htb]
  \centering
  \includegraphics[width=0.83\linewidth]{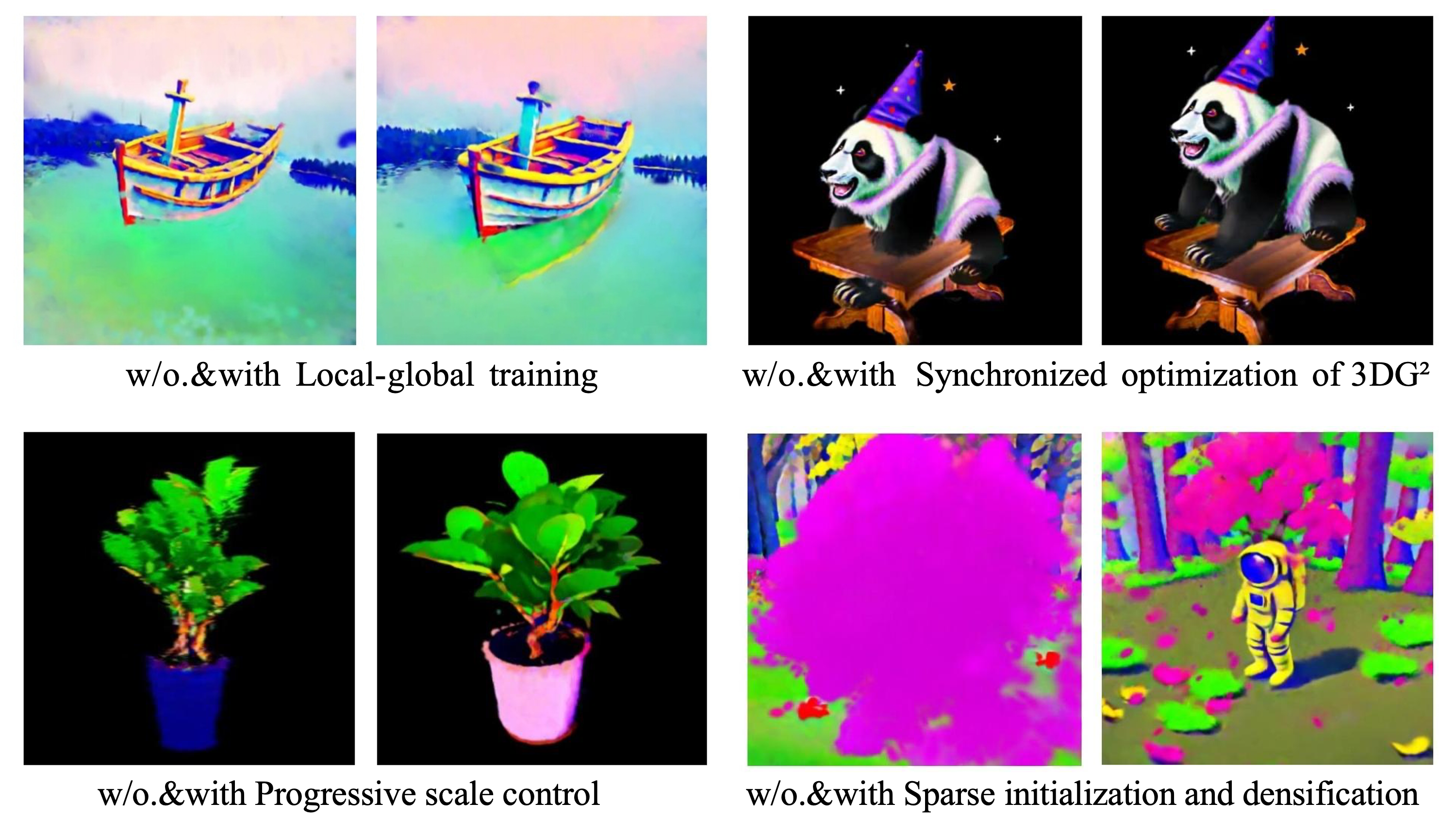}
  \caption{Visualization of ablation experiments. We have carried out ablation experiments on several modules in DreamScape and proved their effectiveness.}
  \label{fi:ablation}
\end{figure}

To validate the effectiveness of our proposed modules, we conducted ablation experiments on local-global training, progressive scale control, synchronized optimization, and sparse initialization with densification. The results, shown in Figure \ref{fi:ablation}, highlight key findings: without local-global training, interactions between objects are absent; synchronized optimization resolves object collisions and misalignments; progressive scale control ensures proper object deformation; and sparse initialization with densification accurately models pervasive objects. These experiments demonstrate the efficacy of our optimization modules in producing higher-quality results.
Additionally, we conducted a user study to further assess generative quality and multi-view consistency. Participants rated 10 randomly selected images generated by different ablated versions of our model on a scale from 1 to 5, resulting in 63 valid responses. The averaged scores, as presented in Table \ref{tab:userstudy}, strongly support the effectiveness and necessity of our proposed modules, demonstrating clear improvements in both visual quality and consistency when incorporating our full optimization framework.
\vspace{-0.3cm}
\begin{figure}[ht]
  \centering
  \includegraphics[width=0.8\linewidth]{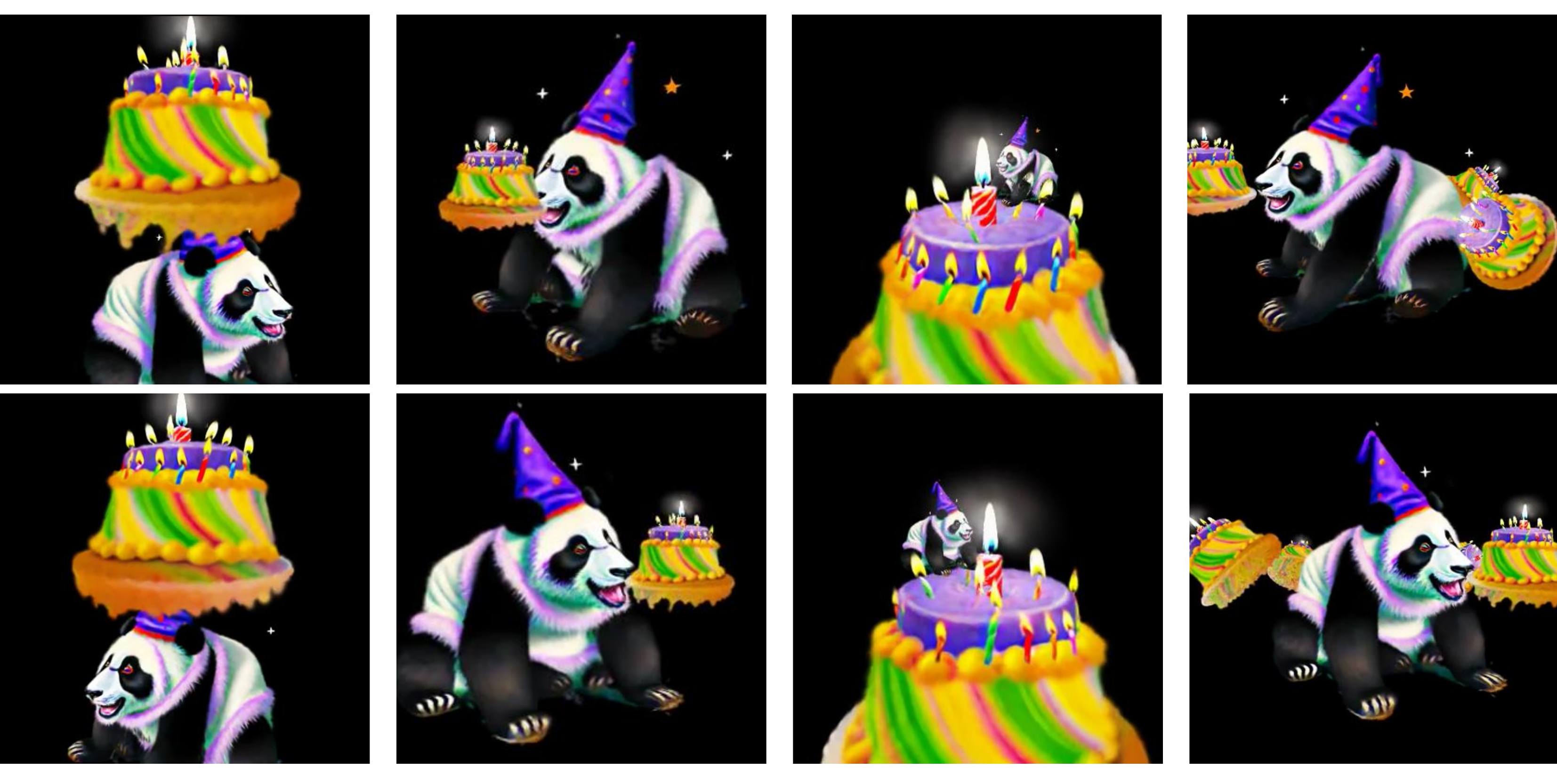}
  \caption{Demonstration of editing ability. DreamScape can edit the generated results in real time, including position transformation, scaling, rotation, etc.}
  \label{fi:edit}
\end{figure}
\subsection{Editing}

As an explicit representation method, 3D Gaussians offer high editability. Unlike methods that treat scenes as a whole \cite{text2room,text2nerf,luciddreamer}, DreamScape decomposes scenes into individual objects for modeling. Users can freely modify the positions, scales, and rotation angles of each object without further inference. This open editing approach allows for convenient application in various practical 3D modeling scenarios. However, it is worth noting that for cases involving interaction effects such as water ripples and light shadows, further optimization may be required after editing to maintain high fidelity.

\section{Conclusion}

In this paper, we propose DreamScape, a novel pipeline for 3D scene generation, which leverages 3D Gaussian Guide as a bridge to facilitate the interaction between LLMs and Diffusion priors using only text descriptions. Through strategies such as local-global training, progressive scale control, and synchronized optimization of the $3{DG^2}$, DreamScape facilitates interaction among multiple objects in the scene, achieving the generation of interactive 3D Gaussian scenes. Furthermore, to address challenges such as difficulty in generating pervasive objects, we proposed sparse initialization and densification strategies, further enhancing the immersion and the atmospheric quality of the generated scenes. 
Despite this, the capabilities of our model are still constrained by the abilities of single-object generation models. 
To address such issues, it is necessary to enhance the capabilities of the these models and tackle biases in the datasets.

\section*{Acknowledgment}
This work is partly supported by the National Key R\&D Program of China (2022ZD0161902), and  the National Natural Science Foundation of China (No. 62202031).  

% \section*{References}

% Please number citations consecutively within brackets \cite{b1}. The 
% sentence punctuation follows the bracket \cite{b2}. Refer simply to the reference 
% number, as in \cite{b3}---do not use ``Ref. \cite{b3}'' or ``reference \cite{b3}'' except at 
% the beginning of a sentence: ``Reference \cite{b3} was the first $\ldots$''

% Number footnotes separately in superscripts. Place the actual footnote at 
% the bottom of the column in which it was cited. Do not put footnotes in the 
% abstract or reference list. Use letters for table footnotes.

% Unless there are six authors or more give all authors' names; do not use 
% ``et al.''. Papers that have not been published, even if they have been 
% submitted for publication, should be cited as ``unpublished'' \cite{b4}. Papers 
% that have been accepted for publication should be cited as ``in press'' \cite{b5}. 
% Capitalize only the first word in a paper title, except for proper nouns and 
% element symbols.

% For papers published in translation journals, please give the English 
% citation first, followed by the original foreign-language citation \cite{b6}.

\vspace{12pt}
\color{red}
% IEEE conference templates contain guidance text for composing and formatting conference papers. Please ensure that all template text is removed from your conference paper prior to submission to the conference. Failure to remove the template text from your paper may result in your paper not being published.

\end{document}